\documentclass[runningheads]{llncs}
\usepackage{graphicx}
\usepackage{marvosym}
\usepackage{pifont}
\usepackage{graphicx}
\usepackage{amsmath}
\usepackage{amssymb}

\DeclareSymbolFont{matha}{OML}{txmi}{m}{it}
\DeclareMathSymbol{\varv}{\mathord}{matha}{118}
\usepackage{array}
\usepackage{tabularx}
\usepackage{multirow, booktabs}
\usepackage{float}
\usepackage{bm}
\usepackage{subfigure}
\usepackage[breaklinks=false,bookmarks=false]{hyperref}
\begin{document}
\title{SAMConvex: Fast Discrete Optimization for CT Registration using Self-supervised Anatomical Embedding and Correlation Pyramid}
%
%
\author{Zi Li\thanks{Equal contribution. \textrm{\Letter} Corresponding author.} \inst{1,4}$^{(\textrm{\Letter})}$ 
\and Lin Tian$^{\star}$\inst{1,2} 
\and Tony C. W. Mok\inst{1,4}
\and Xiaoyu Bai\inst{1,4}
\and Puyang Wang\inst{1,4}
\and Jia Ge\inst{3}
\and Jingren Zhou\inst{1,4}
\and Le Lu\inst{1}
\and Xianghua Ye\inst{3}
\and Ke Yan\inst{1,4}
\and Dakai Jin\inst{1}
}
\authorrunning{Z. Li et al.}
\titlerunning{SAMConvex: Fast Discrete Optimization for CT Registration}
%
\institute{DAMO Academy, Alibaba Group \and
University of North Carolina at Chapel Hill, USA  \and
The First Affiliated Hospital of College of Medicine, Zhejiang University, China \and Hupan Lab, 310023, Hangzhou, China  \\
\email{alisonbrielee@gmail.com; lintian@cs.unc.edu}}
\maketitle              
\begin{abstract}
Estimating displacement vector field via a cost volume computed in the feature space has shown great success in image registration, but it suffers excessive computation burdens. Moreover, existing feature descriptors only extract local features incapable of representing the global semantic information, which is especially important for solving large transformations. 
To address the discussed issues, we propose SAMConvex, a fast coarse-to-fine discrete optimization method for CT registration that includes a decoupled convex optimization procedure to obtain deformation fields based on a self-supervised anatomical embedding (SAM) feature extractor that captures both local and global information.
To be specific, SAMConvex extracts per-voxel features and builds 6D correlation volumes based on SAM features, and iteratively updates a flow field by performing lookups on the correlation volumes with a coarse-to-fine scheme.
SAMConvex outperforms the state-of-the-art learning-based methods and optimization-based methods over two inter-patient registration datasets (Abdomen CT and HeadNeck CT) and one intra-patient registration dataset (Lung CT). Moreover, as an optimization-based method, SAMConvex only takes $\sim2$s ($\sim5s$ with instance optimization) for one paired images.

\keywords{Medical Image Registration \and Large Deformation.}
\end{abstract}
\section{Introduction}\label{sec:intro}
Deformable image registration~\cite{SotirasDP13}, a fundamental medical image analysis task, has traditionally been approached as a continuous optimization~\cite{shen2002hammer,Ashburner07,avants2008symmetric,HeinrichJBS12,SunNK14} problem over the space of dense displacement fields between image pairs.
The iterative process always leads to inefficiency.
Recent learning-based approaches that use a deep network to predict a displacement field~\cite{BalakrishnanZSG19,zhao2019unsupervised,MokC20,LiuLFZHL22,tian2022gradicon,ijcai2020p101,fan2022automated}, yield much faster runtime and have gained huge attention. 
However, they often struggle with versatile applicability due to the fact that they require training per registration task. Moreover, gathering enough data for the training is not a trivial task in practice. In addition, both optimization and learning-based methods rely on similarity measures computed over the intensity, which prevents the methods to utilize the anatomy correspondence. Several works~\cite{HeinrichJBMGBS12,HeinrichJPBS13} use feature descriptors that provide modality and contrast invariant information but they still can only represent local information and do not contain the global semantic information. Thus, they face challenges in settings with large deformations or complex anatomical differences (\emph{e.g.}, inter-patient Abdomen).

To address this issue, we incorporate a Self-supervised Anatomical eMbedding (SAM)~\cite{yan2022sam} into registration. SAM generates a unique semantic embedding for each voxel in CT that describes its anatomical structure, thus, providing semantically coherent information suitable for registration. 
SAME~\cite{LiuYHGLYHXXYJ21} enhances learning-based registration with SAM embeddings, but it suffers the applicability issue as the other learning-based methods even when the SAM embedding is pre-trained and ready to use out of the box for multiple anatomical regions.

Registration has also been formulated as a discrete optimization problem \cite{5459364,Chen_2016_CVPR,Xu2017AccurateOF,Heinrich2019ClosingTG,SiebertHH21} that employs a dense set of discrete displacements, called cost volume. The main challenge of this category of approach is the massive size of the search space, as millions of voxels exist in a typical 3D CT scan and each voxel in the moving scan can be reasonably paired with thousands of points in the other scan, leading to a high computational burden. To obtain fast registration with discrete optimization, Heinrich \emph{et al.}~\cite{heinrich2014non} prunes the search space by constructing the cost volume only within the neighborhood of each voxel. However, the magnitude range of the deformation it can solve is limited by the size of the neighborhood window, leading to reliance on an accurate pre-alignment.  

We propose SAMConvex, a coarse-to-fine discrete optimization method for CT registration. Specifically, SAMConvex presents two main components: (1) a \emph{discriminative} feature extractor that encodes global and local embeddings for each voxel; (2) a \emph{lightweight} correlation pyramid that constructs multi-scale 6D cost volume by taking the inner product of SAM embeddings. It enjoys the strengths of state-of-the-art accuracy (Sec.~\ref{sec:3d_dataset_result}), good generalization (Sec.~\ref{sec:3d_dataset_result} and Sec.~\ref{sec:robustness}) and fast runtime (Sec.~\ref{sec:3d_dataset_result}).

\section{SAMConvex Method}
\subsection{Problem Formulation}
Given a source image $I_s:\Omega_s\to{\mathbb{R}}$ and a target image $I_t:\Omega_t\to{\mathbb{R}}$ within a spatial domain $\Omega\subset{\mathbb{R}^n}$, our goal is to find the spatial transformation $\varphi^{-1}:\Omega_t\to\Omega_s$. We aim at minimizing the following energy:
\begin{equation}
    \hat{u} =  \arg\min_{u}E_\mathcal{D}(I_s\circ\varphi^{-1},I_t)
     + {\lambda}E_\mathcal{R}(u),
\label{eq:regformulation}
\end{equation}
where the transformation model is the displacement vector field (DVF) $\varphi^{-1}=Id+u$ with $Id$ being the identity transformation, the $E_\mathcal{D}$ is the similarity term measuring the similarity between $I_t$ and warped image $I_s\circ\varphi^{-1}$, $E_\mathcal{R}=||\nabla{u}||_2^2$ is the diffusion regularizer that advocates the regularity of $\varphi^{-1}$. $\lambda$ weights between the data matching term and the regularization term.

\begin{figure}[!t]
    \centering
    \includegraphics[width=1.0\textwidth]{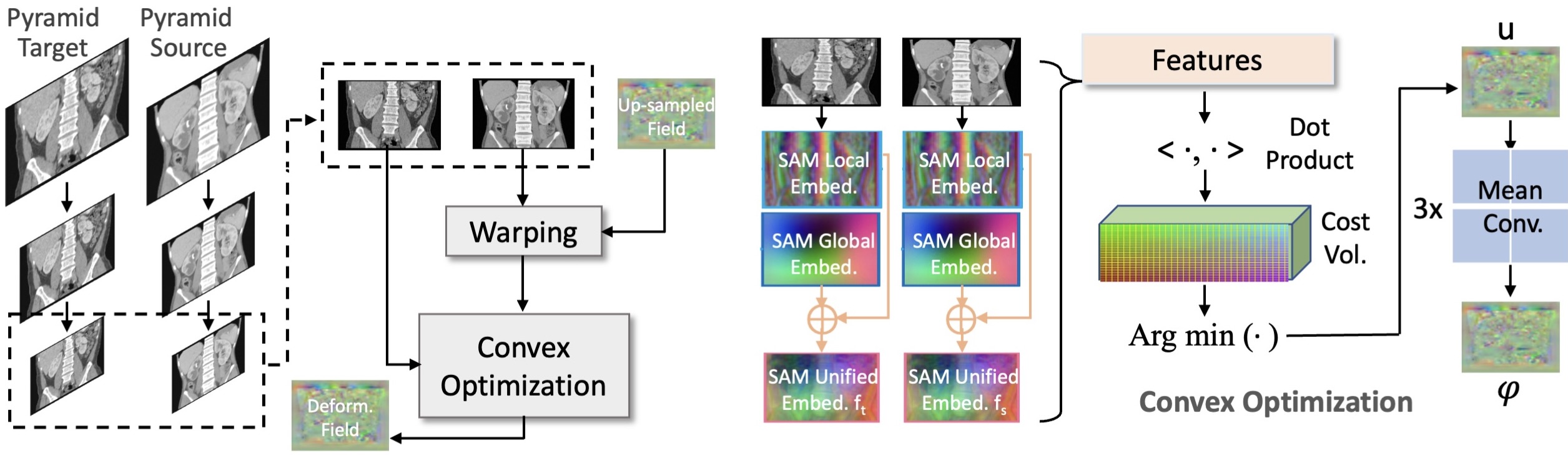}
    \caption{
    \textbf{Left}: Image pyramid and refinement at one pyramid level by discrete registration with convex optimization. \textbf{Right}: the detail of convex optimization that consists of SA pyramid and hierarchical optimization at one pyramid level. }
    \label{fig:pipline}
\end{figure}

\subsection{Decoupling to Convex Optimizations}
We conduct the optimization scheme proposed in ~\cite{chambolle2004algorithm,steinbrucker2009large,zach2007duality,heinrich2014non} and we give a brief review here. By introducing an extra term into Eq.~\ref{eq:regformulation} 
\begin{equation}
    \hat{v}, \hat{u} =  \arg\min_{v,u}E_\mathcal{D}(I_s\circ(Id+v),I_t)
     + \frac{1}{2\theta}\| v - u\|^2 + {\lambda}E_\mathcal{R}(u),
\label{eq:convexoptimization}
\end{equation}
the optimization then can be decomposed into two sub-optimization problems\footnote{For better illustration, we substitute $ {\lambda}E_\mathcal{R}(u)$ and $E_\mathcal{D}(I_s\circ\varphi,I_t)$ with $const$ in the two equations in Eq.~\ref{equ:decoupling}, respectively.} 
\begin{align}
\left\{
 \begin{aligned} \label{equ:decoupling}
\hat{v} = & \arg\min_{v} E_\mathcal{D}(I_s\circ(Id+v),I_t) + \frac{1}{2\theta} \| v - \hat{u}\|^2 + const,\\
\hat{u} =  & \arg\min_{u} \frac{1}{2\theta} \| \hat{v} - u\|^2 +  {\lambda}E_\mathcal{R}(u) + const.
\end{aligned}
\right.
\end{align}
with each can be solved via global optimizations. By alternatively optimizing the two subproblems and progressively reducing $\theta$ during the alternative optimization, we get $\hat{v}\approx\hat{u}$ and obtain the solution of Eq.~\ref{eq:regformulation} as $\varphi^{-1}=Id+\hat{v}$. To be noted, the first problem in Eq.~\ref{equ:decoupling} can be solved point-wise because the spatial derivatives of $v$ is not involved. We can search over the cost volume of each point to obtain the optimal solution of the first problem in Eq.~\ref{equ:decoupling}. For the second problem, we are inspired by mean-field inference~\cite{KrahenbuhlK11} and process $u$ using average pooling. 

\subsection{$E_\mathcal{D}(\cdot, \cdot)$ using Global \& Local Semantic Embedding}
Presumably, handling complex registration tasks rely on: (1) distinct features that are robust to inter-subject variation, organ deformation, contrast injection, and pathological changes, and (2) global/contextual information that can benefit the registration accuracy in complex deformation. 
To achieve these goals, we adopt self-supervised anatomical embedding (SAM)~\cite{yan2022sam} based feature descriptor that encodes both global and local embeddings.
The global embeddings memorize the 3D contextual information of body parts on a coarse level, while the local embeddings differentiate adjacent structures with similar appearances, as shown on the left of Fig.~\ref{fig:pipline}. The former helps the latter to focus on small regions to distinguish with fine-level features.

To be specific, given an image $I \in{\mathbb{R}^{H \times W \times D}}$,  we utilize a pre-trained SAM model that outputs a global embedding $f_{global} \in{\mathbb{R}^{128 \times \frac{H}{8} \times \frac{W}{8} \times \frac{D}{2}}}$ and a local embedding $f_{local} \in{\mathbb{R}^{128 \times \frac{H}{2} \times \frac{W}{2} \times \frac{D}{2}}}$. We resample the global embedding with bilinear interpolation to match the shape of $f_{global}$ to $f_{local}$ and concatenate them to get $f_{SAM} \in{\mathbb{R}^{256 \times \frac{H}{2} \times \frac{W}{2} \times \frac{D}{2}}}$.
We normalize $f_{global}$ and $f_{local}$ before concatenation and adopt the dot product $<\cdot,\cdot>$ as the similarity measure in the following part, with a higher value indicating better alignment. 

With $f_{SAM}$, we construct $E_\mathcal{D}(\cdot, \cdot)$ as the cost volume in the SAM feature space within a neighborhood
of $[-N,N]$, where $N$ represents searching radius. Given two images $I^0$ and $I^1$, the resulting $E_\mathcal{D}(\cdot, \cdot)$ can then be formulated as
\begin{equation}
    E_\mathcal{D}(I^0, I^1) = <f_{SAM}^0(x), f_{SAM}^1(x+d)>
\end{equation}
where $f_{SAM}^0$ and $f_{SAM}^1$ are the SAM embedding of $I^0$ and $I^1$, respectively. $x$ is any voxel in the image domain and $d$ is the voxel displacement within the neighborhood. $E_\mathcal{D}(\cdot, \cdot)$ has the shape of $(2N+1)\times(2N+1)\times(2N+1)\times\frac{H}{2}\times\frac{W}{2}\times\frac{D}{2}$.

\subsection{Coarse-to-fine Optimization Strategy}
Since $E_\mathcal{D}(\cdot, \cdot)$ is computed within the neighborhood, the magnitude of deformation is bounded by the size of the neighborhood. Our approach to addressing this issue is based on a surprisingly simple but effective observation:  performing registration based on a coarse-to-fine scheme with a small search range at each level instead of a large search range at one resolution benefits to 1) significantly improve the efficiency such as low computation burden and fast running time; 2)  enlarge receptive field and improve registration accuracy.

We first build an image pyramid of $I_s$ and $I_t$. At each resolution, we warp the source image with the composed deformation computed from all the previous levels (starting from the coarsest resolution), transform the warped image and target image to the SAM space, and conduct the decoupling to convex optimizations strategy to obtain the deformation between the warped image and target at the current resolution. With such a coarse-to-fine strategy, we estimate a sequence of displacement fields from a starting identity transformation. The final field is computed via the composition of all the displacement fields~\cite{zhao2019unsupervised}. 

To be noted,  the cost volume at each level is computed by taking the inputs of $\{1, \frac{1}{2}, \frac{1}{4}\}$ resolution. Adding one coarser level consumes less computation than doubling the size of the neighborhood at the fine resolution but yields the same search range.

\subsection{Implementation Detail}
We conduct the registration in 3 levels of resolutions. At each level, we solve Eq.~\ref{equ:decoupling} via five alternative optimizations with $ \frac{1}{2\theta} = \{0.003, 0.01, 0.03, 0.1, 0.3, 1\}$.
To further improve the registration performance, we append a SAM-based instance optimization after the coarse-to-fine registration with a learning rate of $0.05$ for $50$ iterations. 
It solves an instance-specific optimization problem~\cite{BalakrishnanZSG19} in the SAM feature space. The optimization objective consists of similarity (dot product between SAM feature vectors on the highest resolution) and diffusion regularization terms.
All the experiments are run on the CPU of Intel Xeon Platinum 8163 with 16 cores of 2.50GHz and the GPU of NVIDIA Tesla V100.
Code will be available at \url{https://github.com/Alison-brie/SAMConvex}.

\section{Experiments}

\subsection{Datasets and Evaluation Metrics}
We split Abdomen and HeadNeck into a training set and test set to accommodate the requirement of the training dataset of comparing learning-based methods. 
Lung dataset contains 35 CT pairs, which is not sufficient for developing learning-based methods. Hence, it is only used as a testing set for optimization-based methods.
All the methods are evaluated on the test set. 
The SAM is pre-trained on NIH Lymph Nodes dataset~\cite{yan2022sam}. All 3 datasets in this paper are not used for pre-training.

\noindent
\textbf{Inter-patient task on Abdomen:}
The Abdomen CT dataset~\cite{hering2022learn2reg} contains 30 abdominal scans with 20 for training and 10 for testing. Each image has 13 manually labeled anatomical structures: spleen, right kidney, left kidney, gall bladder, esophagus, liver, stomach, aorta, inferior vena cava, portal and splenic vein, pancreas, left adrenal gland, and right adrenal gland. 
The images are resampled to the same voxel resolution of 2 mm and spatial dimensions of $192 \times 160 \times 256$. 

\noindent\textbf{Inter-patient task on HeadNeck:}
The HeadNeck CT dataset~\cite{ye2022comprehensive} contains 72 subjects with 13 organs labeled. 
The manually labeled anatomical structures include the brainstem, left eye, right eye, left lens, right lens, optic chiasm, left optic nerve, right optic nerve, left parotid, right parotid, left and right temporomandibular joint, and spinal cord.
We split the dataset into 52, 10, and 10 for training, validation, and test set.
For testing, we construct 90 image pairs for registration, with each image, resampled to an isotropic resolution of 2 mm and cropped to $256 \times 128 \times 224$.

\noindent\textbf{Intra-patient task on Lung:}
The images are acquired as part of the radiotherapy planning process for the treatment of malignancies. 
We collect a lung 4D CT dataset containing 35 patients, each with inspiratory and expiratory breath-hold image pairs, and take this dataset as an extra test set.
Each image has labeled malignancies, and we try to align two phases for motion estimation of malignancies. 
The images are resampled to a spacing of $2\times2\times2$ mm and cropped to $256 \times 256 \times 112$.  All images are used as testing cases.

\noindent\textbf{Evaluation metrics:}
We use the average Dice score (DSC) to evaluate the accuracy and compute the standard deviation of the logarithm of the Jacobian determinant (SDlogJ) to evaluate the plausibility of deformation fields, also comparing running time ($\textnormal{T}_{test}$) on the same hardware. 

\begin{table}[!t]
  \centering
  \caption{Quantitative results of inter- and intra-subject registration. The subscript of each metric indicates the number of anatomical structures involved. $\uparrow$: higher is better, and $\downarrow$: lower is better. Initial: initial affine results without deformable registration. Ours indicates the coarse-to-fine registration and IO indicates the instance optimization.}
  \resizebox{\textwidth}{!}{%
    \begin{tabular}{ccccccccccc}
      \toprule[1.5pt]
      \multirow{2}{*}{Method} & \multicolumn{3}{c}{Abdomen CT} & \multicolumn{3}{c}{HeadNeck CT} & \multicolumn{3}{c}{Lung CT}\\
      \cmidrule(lr){2-4}\cmidrule(lr){5-7}\cmidrule(lr){8-10}
       & \rule{1pt}{0ex} DSC$_{13}$ $\uparrow$ & SDlogJ $\downarrow$ & $\textnormal{T}_{test}$ $\downarrow$ & \rule{1pt}{0ex} DSC$_{13}$ $\uparrow$ & SDlogJ $\downarrow$ & $\textnormal{T}_{test}$ $\downarrow$ & \rule{1pt}{0ex} DSC$_1$ $\uparrow$ & SDlogJ $\downarrow$ & $\textnormal{T}_{test}$ $\downarrow$ \\
      \midrule[1pt]
      Initial \hspace{0.1cm} & 32.64 & - & - & 33.91 & - & - & 67.06 & - & - \\
      \midrule
      NiftyReg & 34.98 & 0.22 & 186.54 s & 57.66 & 0.21 & 168.39 s & 74.59 & 0.02 & 85.50 s \\
      Deeds & 46.52 & 0.44 & 45.21 s & 61.32 & 1.07 & 42.05 s & 81.53 & 0.49 & 227.59 s \\
      ConvexAdam & 44.44 & 0.74 & 6.06 s & 61.45 & 0.77 & 3.12 s & 79.26 & 1.17 & 2.20 s \\
      LapIRN & 46.44 & 0.72 & 0.07 s & 60.16 & 0.58 & 0.03 s & - & - & - \\
      SAME & 47.12 & 0.90 & 6.88 s & 61.35 & 1.00 & 3.34 s & - & - & - \\
      \midrule
      Ours & 48.30 & 0.86 & 2.12 s & 61.88 & 0.79 & 1.88 s & 80.13 & 0.37 & 1.86 s \\
      Ours + IO & \textbf{51.17} & 0.86 & 5.14 s & \textbf{63.72} & 0.88 & 4.59 s & \textbf{81.61} & 0.21 & 4.34 s \\
      \bottomrule[1.5pt]
    \end{tabular}
  }
  \label{tab:main_result}
\end{table}

\begin{figure}[!t]
    \centering
    \includegraphics[width=1.0\textwidth]{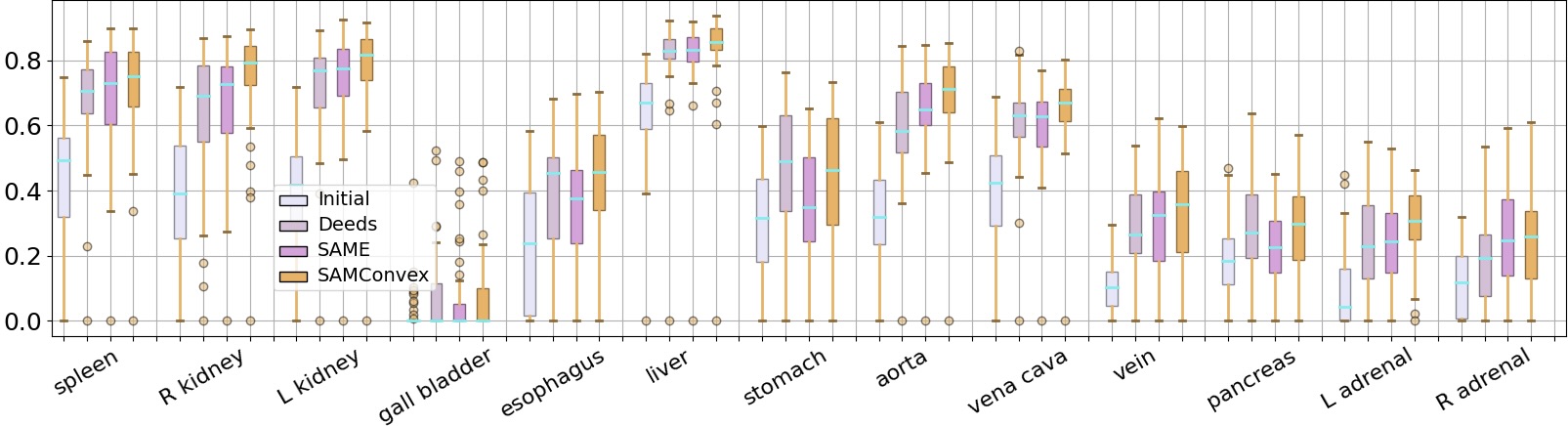}
    \caption{Comparison of performance-top-three deformable registration methods on all organ groups on \textbf{Abdomen} dataset.}
    \label{fig:boxplot}
\end{figure}

\subsection{Registration Results}\label{sec:3d_dataset_result}
We compare with five widely-used and top-performing deformable registration methods, including NiftyReg~\cite{SunNK14}, Deeds~\cite{HeinrichJBS12}, top-ranked ConvexAdam~\cite{SiebertHH21} in Learn2Reg Challenge~\cite{hering2022learn2reg}, and SOTA learning-based methods LapIRN~\cite{MokC20} and SAME~\cite{LiuYHGLYHXXYJ21}.  All results are based on the SAM-affine~\cite{LiuYHGLYHXXYJ21} pre-alignment.

As shown in Table.~\ref{tab:main_result}, SAMConvex outperforms the widely-used NiftyReg~\cite{SunNK14} across the three datasets with an average of \boldsymbol{$10.0\%$} Dice score improvement. 
Compared with the best traditional optimization-based method Deeds~\cite{HeinrichJBS12}, SAMConvex performs better or comparatively on the three datasets with approximately \boldsymbol{$20\sim100$} times faster runtime. 
We also see a better performance of SAMConvex over ConvexAdam~\cite{SiebertHH21}. To be noted, the performance gap between SAMConvex and ConvexAdam is greater on Abdomen CT than the other two datasets. This may be because the deformation is more complex in Abdomen and the anatomical differences between inter-subjects make the registration more challenging. With a coarse-to-fine strategy, SAMConvex can better bear these issues. 
Although LapIRN~\cite{MokC20} has the fastest inference time, it has slightly inferior registration accuracy as compared to SAMConvex. Moreover, it needs training when being applied to a new dataset.
Equipped with SAM, SAME achieves the overall 2nd best performance in two inter-patient registration tasks but with a notably higher folding rate overall. Moreover, it also requires individual training for a new dataset and struggles with the intra-patient lung registration task (small dataset and no available training data).
In summary, our SAMConvex achieves the overall best performance as compared to other methods with a fast inference time and comparable deformation field smoothness.

To better understand the performance of different deformable registration methods, we display organ-specific results of the inter-patient registration task of abdomen CT in Fig.~\ref{fig:boxplot} where large deformations and complex anatomical differences exist.
As shown, SAMConvex is consistently better than the second-best and third-best registration methods on most examined abdominal organs, in 11 out of 13 organs.  

\subsection{Ablation Study on SAMConvex}

\begin{figure}[!t]
	\centering
	\begin{tabular}{c@{\extracolsep{0.1em}}c@{\extracolsep{0.1em}}c@{\extracolsep{0.1em}}c}	
	    \includegraphics[width=0.245\textwidth]{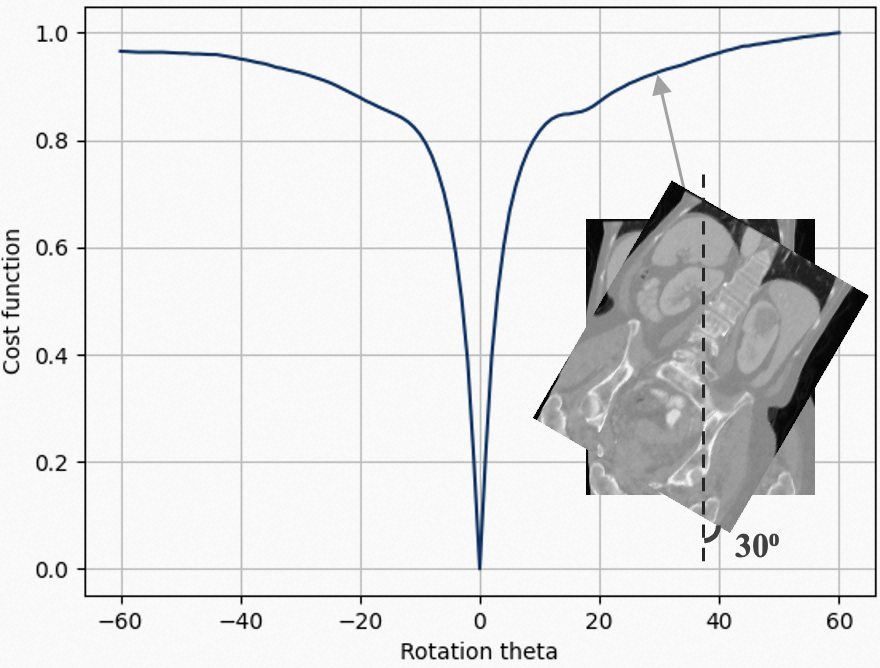}
		&\includegraphics[width=0.245\textwidth]{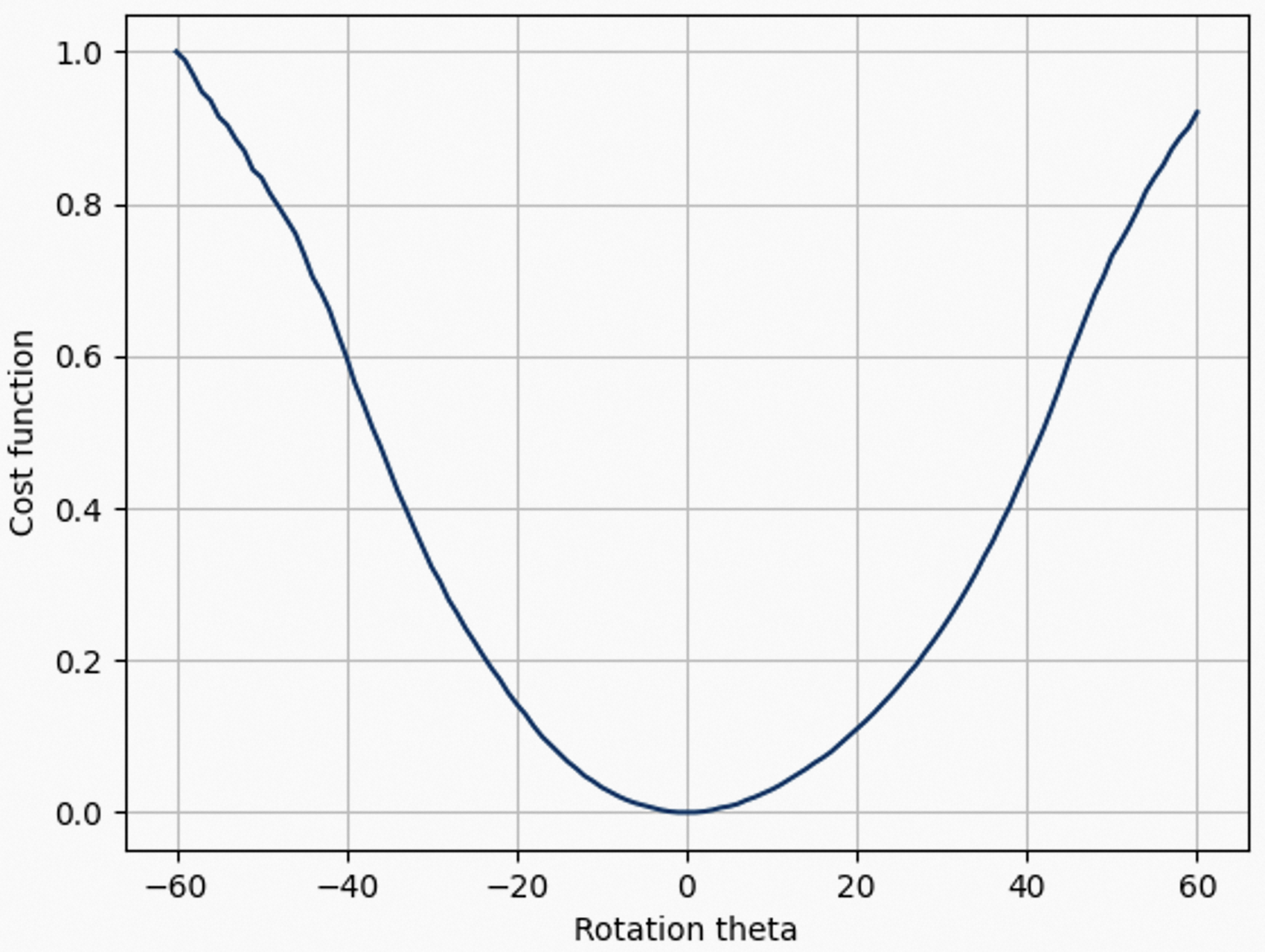}
		&\includegraphics[width=0.245\textwidth]{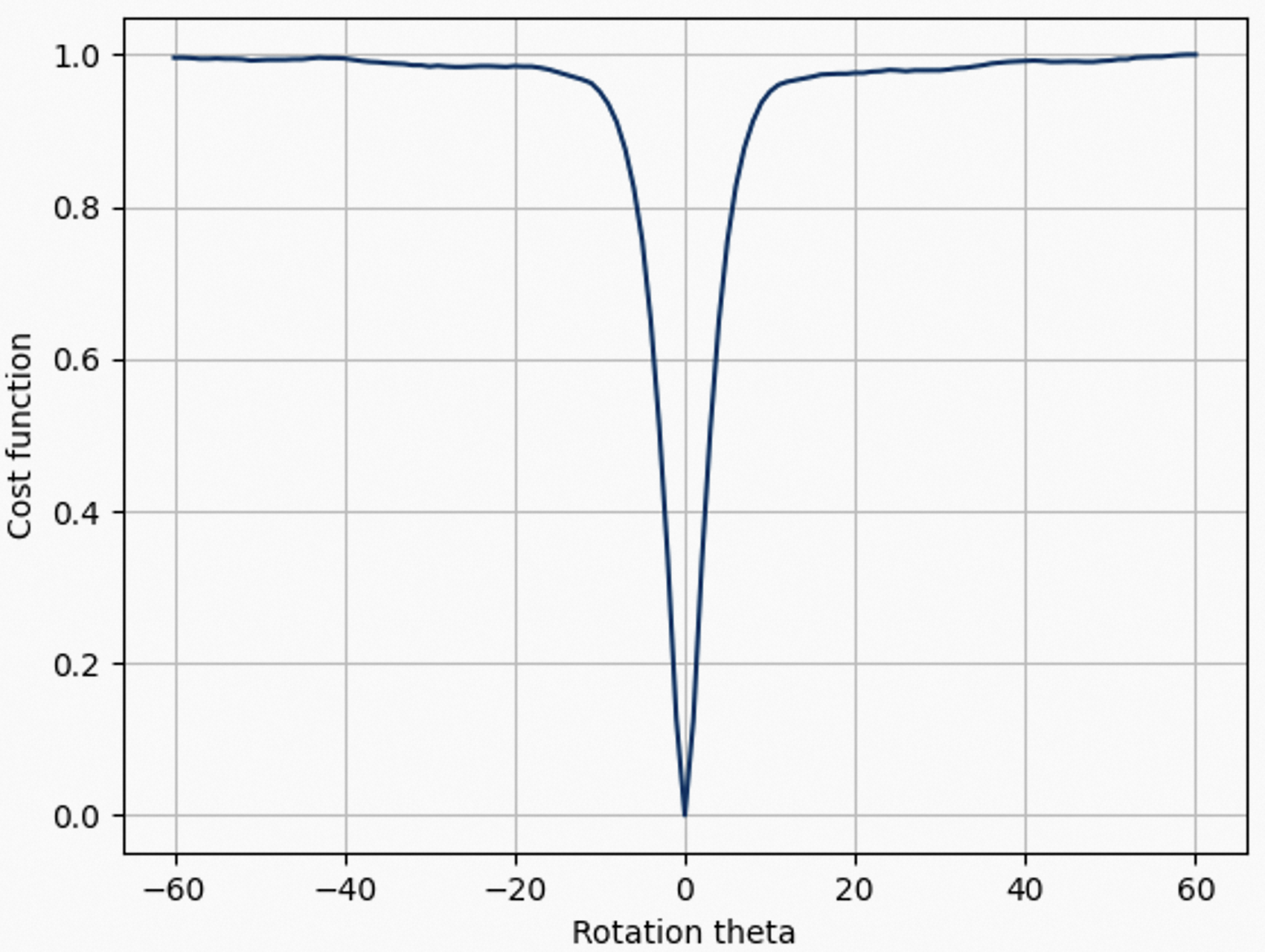}	&\includegraphics[width=0.245\textwidth]{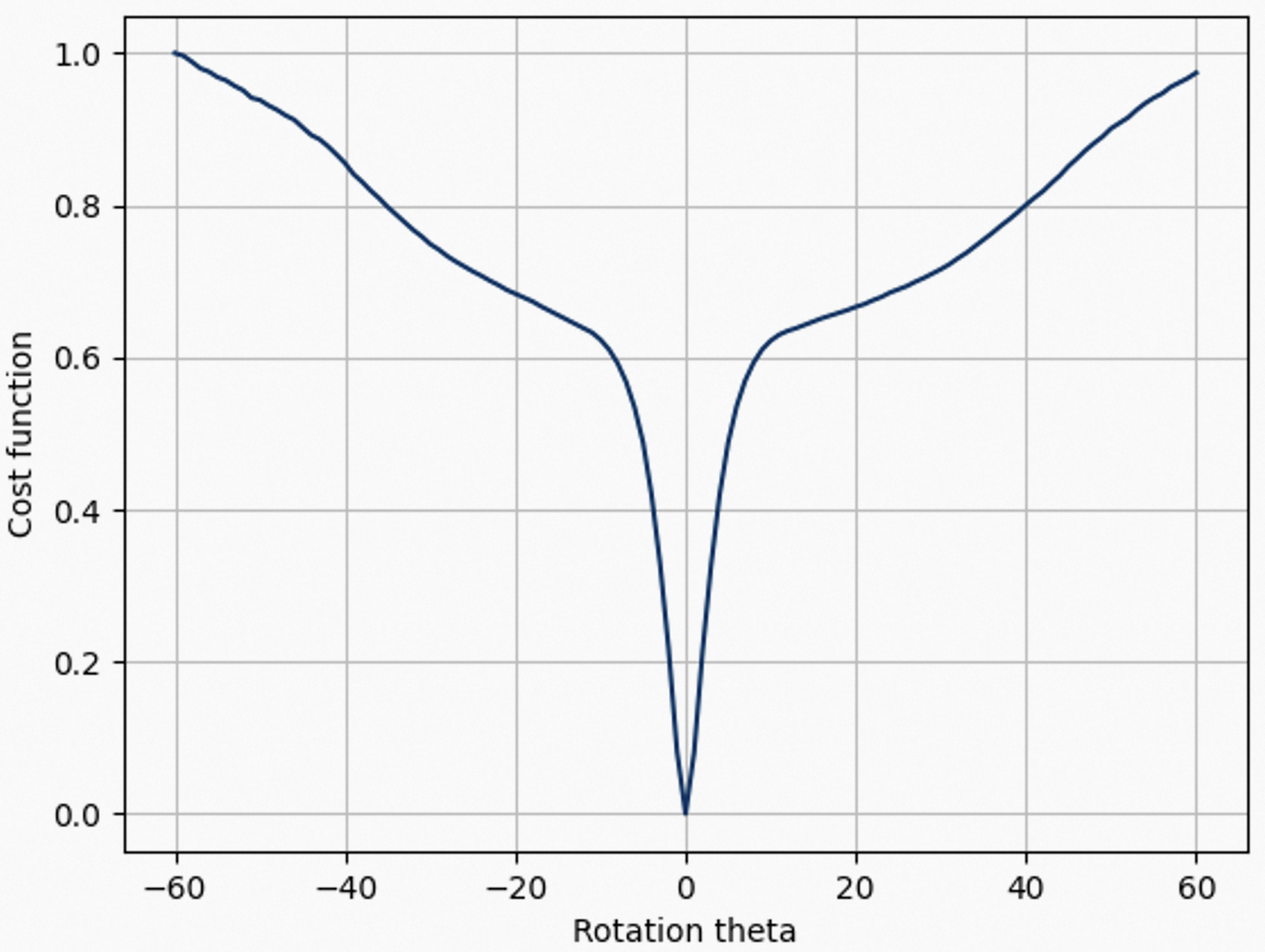}\\
		\includegraphics[width=0.235\textwidth]{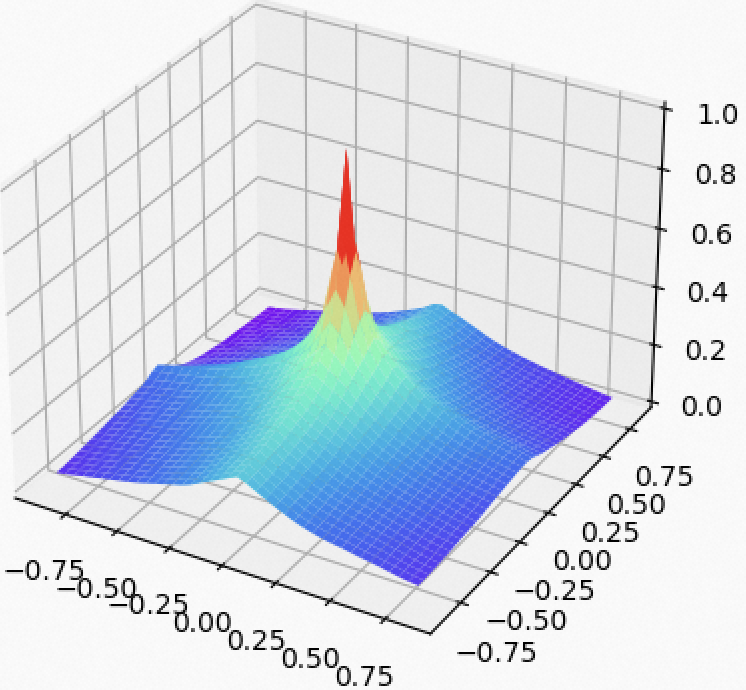}
		&\includegraphics[width=0.235\textwidth]{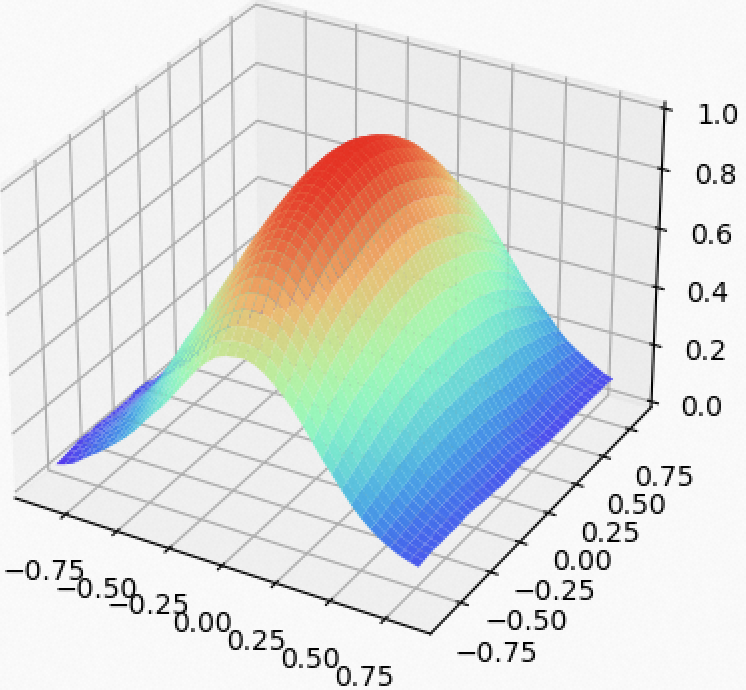}
		&\includegraphics[width=0.235\textwidth]{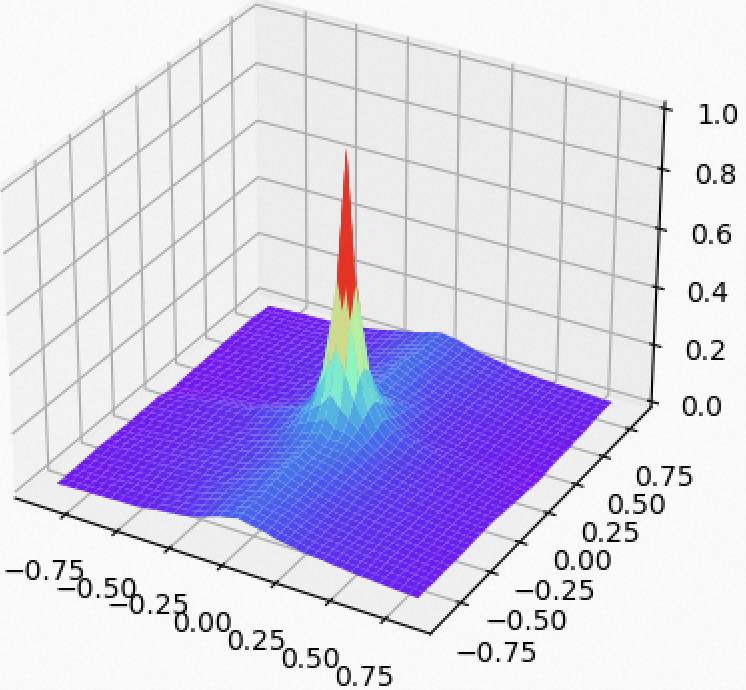}	&\includegraphics[width=0.235\textwidth]{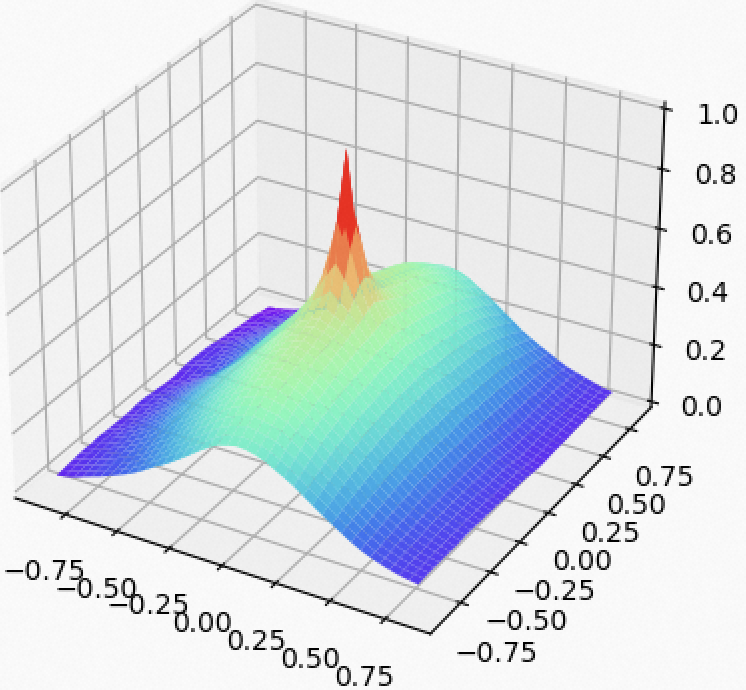}\\
		(a) MIND & (b) SAM global & (c) SAM local & (d) SAM
	\end{tabular}
	\caption{ Loss landscapes of MIND, SAM-global, SAM-local and SAM on \textbf{Abdomen} dataset with varying rotations along two axes.}
	\label{fig:featureAb} 
\end{figure}

\paragraph{Loss landscape of SAM} We explore the loss landscapes of SAM-global, SAM-local, SAM (SAM-global and SAM-local), and MIND via varying the transformation parameters. 
We conduct experiments on the abdomen dataset~\footnote{Refer to Appendix for more experiment results.}. Fig.~\ref{fig:featureAb} shows the comparison of landscapes when we vary the rotation along two axes. The loss landscape falls to flat quickly when the rotation is greater than $~20^{\circ}$ degrees and $~40^{\circ}$ degrees for SAM-local and MIND, respectively. The flattened area indicates where the similarity measure loses the capability to guide the registration because multiple transformation parameters yield the same similarity value. Compared with SAM-local and MIND, SAM-global does not have the flattened area within $[-60^{\circ}, 60^{\circ}]$, meaning it can give correct measure over a larger capture range. However, it does not have the fast converging area shown as a spike for the loss landscapes of MIND and SAM-local. When combining SAM-global and SAM-local together, SAM shows a greater capture range meanwhile fast convergence when the transformation is small.

\paragraph{Robustness to pre-alignment.}\label{sec:robustness} We study the robustness of SAMConvex over different affine pre-alignment. Elastix-Affine~\cite{KleinSMVP10} and SAM-Affine are used as the pre-alignment and results are notated as Initial$_e$ and Initial$_a$ in Tab.~\ref{tab:featureInit}, respectively. Compared to ConvexAdam, our method SAMConvex is less affected by the performance of the Affine pre-alignment. This is in line with our expectations. The global information contained in SAM provides $E_\mathcal{D}(I_s\circ\varphi,I_t)$ a greater capture range than features that contain local information only. Thus, SAMConvex can register images with larger transformation, leading to less reliance on the performance of the pre-alignment.

\paragraph{Ablation on coarse-to-fine.} We study how the number of coarse-to-fine layers and how the size of the neighborhood window affect the registration result on the Abdomen dataset. From Tab.~\ref{tab:pym}, we can conclude that performing registration with a small search range with a coarse-to-fine scheme instead of a cost volume with a large search range can help to improve registration accuracy and computation efficiency.

\begin{minipage}{\textwidth}
\begin{minipage}[t]{0.35\textwidth}
\makeatletter\def\@captype{table}
\small
\centering
\caption{ConvexAdam and SAMConvex on different affine pre-alignment.}
\begin{tabular}{ccc}
    \toprule
    \multirow{2}{*}{Method}  & \multicolumn{2}{c}{DSC$_{13}$ $\uparrow$}  \\
    \cmidrule(lr){2-3} 
      &  Initial$_{e}$ & Initial$_{a}$ \\
    \midrule
    Initial    & 25.88 &  32.64 \\
    ConvexAdam   & 34.42 & 44.44\\
    SAMConvex  & 45.38 & 48.30  \\
    \bottomrule
    \end{tabular}
\label{tab:featureInit}
\end{minipage}
\begin{minipage}[t]{0.55\textwidth}
\makeatletter\def\@captype{table}
\small
\centering
\caption{Ablation study on how different pyramid designs affect the performance. The subscript of Pym indicates the number of scales.}
\begin{tabular}{cccc}
    \toprule
    Method
       & DSC$_{13}$ $\uparrow$ & SDlogJ $\downarrow$ & $\textnormal{T}_{test}$ $\downarrow$ \\
    \midrule
    Pym$_{1}$ $ N \in [3] $     & 40.51     & 0.59      & 1.69 s \\
    Pym$_{1}$ $N \in [6]$    & 44.22     & 1.62     & 7.84 s \\
    Pym$_{2}$ $N \in [3, 3]$  & 47.80     & 0.76     & 2.03 s \\
    Pym$_{3}$ $N \in [2, 3, 3]$   & 48.30 & 0.86 & 2.12 s \\
    \bottomrule
    \end{tabular}
\label{tab:pym}
\end{minipage}
\end{minipage}

\paragraph{Discussion about the differences with ConvexAdam~\cite{SiebertHH21}.} Apart from the SAM extraction module, first, we introduce a cost volume pyramid to the convex optimization framework to reduce the intensive computation burdens. To be specific, with the coarse-to-fine strategy, one can sparsely search on a coarser level with a smaller search radius in each iteration, reaching the same search range as the complete search with less computational complexity. Second, we explicitly validate the robustness of the SAM feature against geometrical transformations and integrate the SAM feature into an instance-specific optimization pipeline. Our SAMConvex is less sensitive to local minimal, achieving superior performance with comparable running time.
Ablation study on pyramid designs (Tab.~\ref{tab:pym}) and leading accuracy on large deformation registration tasks (Tab.~\ref{tab:main_result}) further support our claims.

\section{Conclusion}
We present SAMConvex, a coarse-to-fine discrete optimization method for CT registration. It extracts per-voxel semantic global and local features and builds a series of lightweight 6D correlation volumes, and iteratively updates a flow field by performing lookups on the correlation volumes.
The performance on two inter-patient and one intra-patient registration datasets demonstrates state-of-the-art accuracy, good generalization, and high computation efficiency of SAMConvex.

\bibliographystyle{splncs04}
\bibliography{refs}

\end{document}